\documentclass[10pt,twocolumn,letterpaper]{article}

\usepackage{cvpr}
\usepackage{times}
\usepackage{graphicx}
\usepackage{subfig}
\graphicspath{{./figures/}}
\usepackage{amsmath}
\usepackage{amssymb}
\usepackage{leftidx}
\usepackage{multirow}
\usepackage{makecell}
\usepackage{pgfplots}
\pgfplotsset{compat=1.5}
\usepackage{tikz}
\usepackage{bm}

\usepackage{booktabs}

\usepackage[pagebackref=true,breaklinks=true,colorlinks,bookmarks=false]{hyperref}

\cvprfinalcopy 

\def\cvprPaperID{2499} 

\ifcvprfinal\fi
\begin{document}

\title{Mobile Video Object Detection with Temporally-Aware Feature Maps}

\author{Mason Liu\\
Georgia Tech\thanks{This work was done while interning at Google.} \\
\texttt{\small masonliuw{@}gatech.edu}\\
\and
Menglong Zhu\\
Google\\
\texttt{\small menglong{@}google.com}\\
}

\maketitle
\thispagestyle{empty}

\begin{abstract}
    This paper introduces an online model for object detection in videos designed to run in real-time on low-powered mobile and embedded devices. Our approach combines fast single-image object detection with convolutional long short term memory (LSTM) layers to create an interweaved recurrent-convolutional architecture. Additionally, we propose an efficient Bottleneck-LSTM layer that significantly reduces computational cost compared to regular LSTMs. Our network achieves temporal awareness by using Bottleneck-LSTMs to refine and propagate feature maps across frames. This approach is substantially faster than existing detection methods in video, outperforming the fastest single-frame models in model size and computational cost while attaining accuracy comparable to much more expensive single-frame models on the Imagenet VID 2015 dataset. Our model reaches a real-time inference speed of up to 15 FPS on a mobile CPU.
\end{abstract}

\section{Introduction}
Convolutional neural networks \cite{Krizhevsky, Simonyan1, Szegedy1, he2016deep} have been firmly established as the state-of-the-art in single-image object detection \cite{Girshick, Hek3, RenS, liu2016ssd, DaiJ}. However, the large memory overhead and slow computation time of these networks have limited their practical applications. In particular, efficiency is a primary consideration when designing models for mobile and embedded platforms. Recently, new architectures \cite{Howard, ZhangX} have allowed neural networks to run on low computational budgets with competitive performance in single-image object detection. However, the video domain presents the additional opportunity and challenge of leveraging temporal cues, and it has been unclear how to make a comparably efficient detection framework for video scenes. This paper investigates the idea of building upon these single-frame models by adding temporal awareness while preserving their speed and low resource consumption.

\begin{figure}
\centering
\subfloat{\includegraphics[width=.3\textwidth]{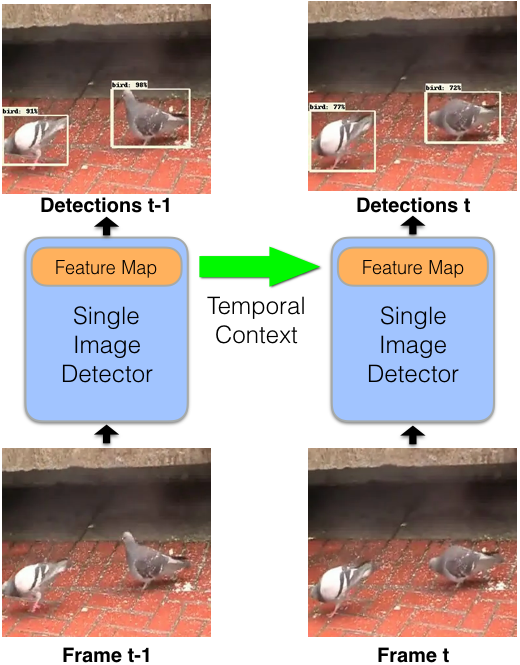}}\\
\caption{Most recent video object detection methods still partially rely on single-image detectors, meaning that a key step in the process fails to utilize temporal information. Our work augments these base detectors by directly incorporating temporal context without sacrificing the efficiency of the fastest single-image detectors.}
\label{fig:3}
\end{figure}

Videos contain various temporal cues which can be exploited to obtain more accurate and stable object detection than in single images. Since videos exhibit temporal continuity, objects in adjacent frames will remain in similar locations and detections will not vary substantially. Hence, detection information from earlier frames can be used to refine predictions at the current frame. For instance, since the network is able to view an object at different poses across frames, it may be able to detect the object more accurately. The network will also become more confident about predictions over time, reducing the problem of instability in single-image object detection \cite{szegedy2013intriguing}.

Recent work has shown that this continuity extends into the feature space, and intermediate feature maps extracted from neighboring frames of video are also highly correlated \cite{ZhuX1}. In our work, we are interested in adding temporal awareness in the feature space as opposed to only on the final detections due to the greater quantity of information available in intermediate layers. We exploit continuity at the feature level by conditioning the feature maps of each frame on corresponding feature maps from previous frames via recurrent network architectures.

Our method generates feature maps with a joint convolutional recurrent unit formed by combining a standard convolutional layer with a convolutional LSTM. The goal is for the convolutional layer to output a feature map hypothesis, which is then fed into the LSTM and fused with temporal context from previous frames to output a refined, temporally aware feature map. Figure \ref{fig:overview} shows an illustration of our method. This approach allows us to benefit from advances in efficient still-image object detection, as we can simply extend some convolutional layers in these models with our convolutional-recurrent units. The recurrent layers are able to propagate temporal cues across frames, allowing the network to access a progressively greater quantity of information as it processes the video. 

To demonstrate the effectiveness of our model, we evaluate on the Imagenet VID 2015 dataset \cite{Russakovsky}. Our method compares favorably to efficient single-frame baselines, and we argue that this improvement must be due to successful utilization of temporal context since we are not adding any additional discriminative power. We present model variants that range from 200M to 1100M multiply-adds and 1 to 3.5 million parameters, which are straightforward to deploy on a wide variety of mobile and embedded platforms. To our knowledge, our model is the first mobile-first neural network for video object detection.

The contributions of this paper are as follows:
\begin{itemize}
  \itemsep0em 
  \item We introduce a unified architecture for performing online object detection in videos that requires minimal computational resources.
  \item We propose a novel role for recurrent layers in convolutional architectures as temporal refinement of feature maps.
  \item We demonstrate the viability of using recurrent layers in efficiency-focused networks by modifying convolutional LSTMs to be highly efficient.
  \item We provide experimental results to justify our design decisions and compare our final architecture to various single-image detection frameworks.
\end{itemize}

\section{Related Work}

\subsection{Object Detection in Images}
Recent single-image object detection methods can be grouped into two categories. One category consists of region-based two-stage methods popularized by R-CNN \cite{Girshick} and its descendants \cite{girshick2015fast, RenS, HongS}. These methods perform inference by first proposing object regions and then classifying each region. Alternatively, single-shot approaches \cite{liu2016ssd, Redmon, DaiJ, focalloss} perform inference in a single pass by generating predictions at fixed anchor positions. Our work builds on top of the SSD framework \cite{liu2016ssd} due to its efficiency and competitive accuracy.

\subsection{Video Object Detection with Tracks}
Many existing methods on the Imagenet VID dataset organize single-frame detections into tracks or tubelets and update detections using postprocessing. Seq-NMS \cite{HanW} links high-confidence predictions into sequences across the entire video. TCNN \cite{KangK1, KangK2} uses optical flow to map detections to neighboring frames and suppresses low-confidence predictions while also incorporating tracking algorithms. However, neither of these methods perform online inference, nor do they focus on efficiency.

Instead of operating on final detection results, our approach directly incorporates temporal context on the feature level and requires no postprocessing steps or online learning. Since our method can be extended by applying any of these approaches to our detection results, our network is more comparable to the base single-frame detector used by these methods than the methods themselves.

A recent work, D$\&$T \cite{feichtenhofer2017detect}, combines tracking and detection by adding an RoI tracking operation and multi-task losses on frame pairs. However, their experiments focus on expensive high-accuracy models, whereas our approach is tailored for mobile environments. Even with aggressive temporal striding, their model cannot be run on mobile devices due to the cost of their Resnet-101 base network, which requires 79$\times$ more computation than our larger model and 450$\times$ more than our smaller one. 

\subsection{Video Object Detection with Optical Flow}
A different class of ideas which powered the winning entry of the 2017 Imagenet VID challenge \cite{ZhuX2} involves using optical flow to warp feature maps across neighboring frames. Deep Feature Flow (DFF) \cite{ZhuX1} accelerates detection by running the detector on sparse key frames and using optical flow to generate the remaining feature maps. Flow-guided Feature Aggregation \cite{ZhuX2} improves detection accuracy by warping and averaging features from nearby frames with adaptive weighting.

These ideas are closer to our approach since they also use temporal cues to directly modify network features and save computation. However, these methods require optical flow information, which is difficult to obtain both quickly and accurately. Even the smallest flow network considered by DFF is approximately twice as computationally expensive as our larger model and 10$\times$ more expensive than our smaller one. Additionally, the need to run both the flow network and the very expensive key frame detector in tandem makes this approach require far more memory and storage consumption than ours, which poses a problem on mobile devices. In contrast, our network contains fewer parameters than our already-efficient baseline detector and has no additional memory overhead.

\subsection{LSTMs for Video Analysis}
LSTMs \cite{Hochreiter} have been successfully applied to many tasks involving sequential data \cite{Lipton}. One specific variant, the convolutional LSTM \cite{ShiX, Patraucean}, uses a 3D hidden state and performs gate computations using convolutional layers, allowing the LSTM to encode both spatial and temporal information. Our work modifies the convolutional LSTM to be more efficient and uses it to propagate temporal information across frames.

Other works have also used LSTMs for video-based tasks. LRCNs \cite{Donahue} use convolutional networks to extract features from each frame and feed these features as a sequence into an LSTM network. ROLO \cite{NingG} performs object tracking by first running the YOLO detector \cite{Redmon} on each frame, then feeding the output bounding boxes and final convolutional features into an LSTM network. While these methods apply LSTMs as postprocessing on top of network outputs, our method fully integrates LSTMs into the base convolutional network via direct feature map refinement.

\subsection{Efficient Neural Networks}
Finally, several techniques for creating more efficient neural network models have previously been proposed, such as quantization \cite{WuJ, ZhouA}, factorization \cite{Jaderberg, Lebedev}, and Deep Compression \cite{HanS}. Another option is creating efficient architectures such as Squeezenet \cite{Iandola}, Mobilenet \cite{Howard}, Xception \cite{Chollet}, and Shufflenet\cite{ZhangX}. In the video domain, NoScope \cite{KangD} uses a modified form of distillation \cite{Hinton} to train an extremely lightweight specialized model at the cost of generalization to other videos.
Our work aims to creating a highly efficient architecture in the video domain where the aforementioned methods are also applicable.

\section{Approach}

\begin{figure}
\centering
\subfloat{\includegraphics[width=\columnwidth]{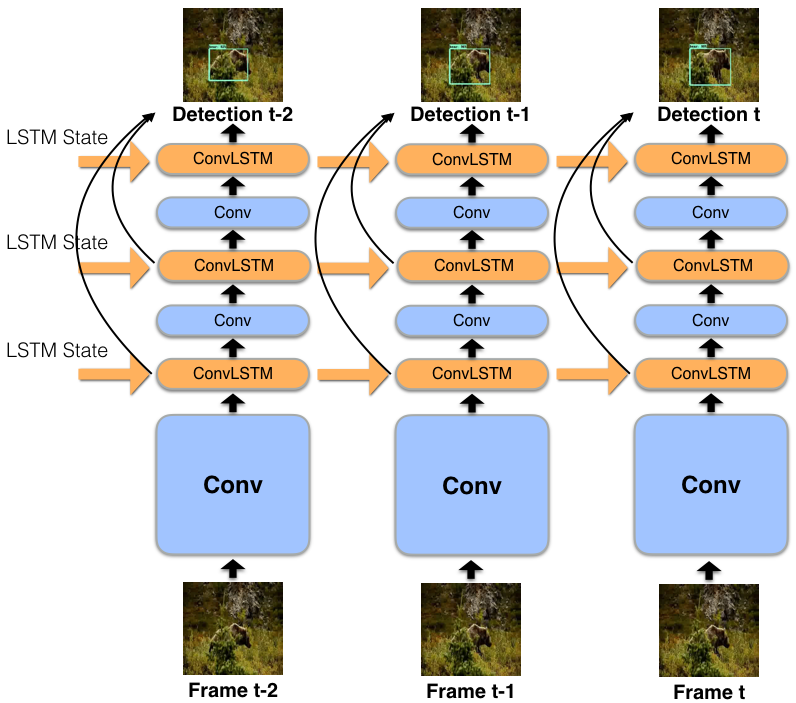}}\\
\caption{An example illustration of our joint LSTM-SSD model. Multiple Convolutional LSTM layers are inserted in the network. Each propagates and refines feature maps at a certain scale.}
\label{fig:overview}
\end{figure}

In this section, we describe our approach for efficient online object detection in videos. At the core of our work, we propose a method to incorporate convolutional LSTMs into single-image detection frameworks as a means of propagating frame-level information across time. However, a naive integration of LSTMs incurs significant computational expenses and prevents the network from running in real-time. To address this problem, we introduce a Bottleneck-LSTM with depthwise separable convolutions \cite{Howard, Chollet} and bottleneck design principles to reduce computational costs. Our final model outperforms analogous single-frame detectors in accuracy, speed, and model size.

\subsection{Integrating Convolutional LSTMs with SSD}

Consider a video as a sequence of image frames $\mathcal{V} = \{I_0, I_1,\ldots I_n\}$. Our goal is to recover frame-level detections $\{D_0, D_1,\ldots D_n\}$, where each $D_k$ is a list of bounding box locations and class predictions corresponding to image $I_k$. Note that we consider an online setting where detections $D_k$ are predicted using only frames up to $I_k$.

Our prediction model can be viewed as a function $\mathcal{F}(I_t, \boldsymbol{s_{t-1}}) = (D_t, \boldsymbol{s_t})$, where $\boldsymbol{s_k}=\{s^0_k, s^1_k,\ldots s^{m-1}_k\}$ is defined as a vector of feature maps describing the video up to frame $k$. We can use a neural network with $m$ LSTM layers to approximate this function, where each feature map of $\boldsymbol{s_{t-1}}$ is used as the state input to one LSTM and each feature map of $\boldsymbol{s_t}$ is retrieved from the LSTM's state output. To obtain detections across the entire video, we simply run each image through the network in sequence.

To construct our model, we first adopt an SSD framework based on the Mobilenet architecture and replace all convolutional layers in the SSD feature layers with depthwise separable convolutions. We also prune the Mobilenet base network by removing the final layer. Instead of having separate detection and LSTM networks, we then inject convolutional LSTM layers directly into our single-frame detector. Convolutional LSTMs allow the network to encode both spatial and temporal information, creating a unified model for processing temporal streams of images.

\subsection{Feature Refinement with LSTMs}
When applied to a video sequence, we can interpret our LSTM states as features representing temporal context. The LSTM can then use temporal context to refine its input at each time step, while also extracting additional temporal cues from the input and updating its state. This mode of refinement is general enough to be applied on any intermediate feature map by placing an LSTM layer immediately after it. The feature map is used as input to the LSTM, and the LSTM's output replaces the feature map in all future computations.

Let us first define our single-frame detector as a function $\bm{G}(I_t) = D_t$. This function will be used to construct a composite network with $m$ LSTM layers. We can view placing these LSTMs as partitioning the layers of $\bm{G}$ into $m+1$ sub-networks $\{g_0, g_1, \ldots g_m\}$ satisfying 
\begin{equation}
\bm{G}(I_t)= (g_m \circ \cdots \circ g_{1} \circ g_{0})(I_t).
\end{equation}
We also define each LSTM layer $\mathcal{L}_0, \mathcal{L}_1, \ldots \mathcal{L}_{m-1}$ as a function $\mathcal{L}_k(M, s^k_{t-1}) = (M_{+}, s^k_t)$, where $M$ and $M_{+}$ are feature maps of the same dimension. Now, by sequentially computing
\begin{align*}
(M^0_{+}, s^0_t) =&\; \mathcal{L}_0(g_0(I_t), s^0_{t-1})\\
(M^1_{+}, s^1_t) =&\; \mathcal{L}_1(g_1(M^0_{+}), s^1_{t-1})\\
\vdots \\
(M^{m-1}_{+}, s^{m-1}_t) =&\; \mathcal{L}_{m-1}(g_{m-1}(M^{m-2}_{+}), s^{m-1}_{t-1})\\
D_t =&\; g_m(M^{m-1}_{+}),
\end{align*}
we have formed the function $\mathcal{F}(I_t, \boldsymbol{s_{t-1}}) = (D_t, \boldsymbol{s_t})$ representing our joint LSTM-SSD network. Figure \ref{fig:overview} depicts the inputs and outputs of our full model as it processes a video. In practice, the inputs and outputs of our LSTM layers can have different dimensions, but the same computation can be performed as long as the first convolutional layer of each sub-network $\mathcal{F}$ has its input dimensions modified.

In our architecture, we choose the partitions of $\bm{G}$ experimentally. Note that placing the LSTM earlier results in larger input volumes, and the computational cost quickly becomes prohibitive. To make the added computation feasible, we consider LSTM placements only after feature maps with the lowest spatial dimensions, which is limited to the Conv13 layer (see Table \ref{tab:1}) and SSD feature maps. In section \hyperref[sec:4.2]{4.2}, we empirically show that placing LSTMs after the Conv13 layer is most effective. Within these constraints, we consider several ways to place LSTM layers:
\begin{enumerate}
  \itemsep0em 
  \item Place a single LSTM after the Conv13 layer.
  \item Stack multiple LSTMs after the Conv13 layer.
  \item Place one LSTM after each feature map
\end{enumerate}

\subsection{Extended Width Multiplier}
LSTMs are inherently expensive due to the need to compute several gates in a single forward pass, which presents a problem in efficiency-focused networks. To address this, we introduce a host of changes that make LSTMs compatible with the goal of real-time mobile object detection.

First, we address the dimensionality of the LSTM. We can obtain finer control over the network architecture by extending the channel width multiplier $\alpha$ defined in \cite{Howard}. The original width multiplier is a hyperparameter used to scale the channel dimension of each layer. Instead of applying this multiplier uniformly to all layers, we introduce three new parameters $\alpha_{base}, \alpha_{ssd}, \alpha_{lstm}$, which control the channel dimensions of different parts of the network.

Any given layer in the base Mobilenet network with $N$ output channels is modified to have $N\alpha_{base}$ output channels, while $\alpha_{ssd}$ applies to all SSD feature maps and $\alpha_{lstm}$ applies to LSTM layers. For our network, we set $\alpha_{base}=\alpha$, $\alpha_{ssd}=0.5\alpha$, and $\alpha_{lstm}=0.25\alpha$. The output of each LSTM is one-fourth the size of the input, which drastically cuts down on the computation required.

\subsection{Efficient Bottleneck-LSTM Layers}
\begin{table}[t]
\resizebox{0.8\columnwidth}{!}{
\scriptsize
\begin{tabular}{c | c | c}
Layer & Filter Size & Stride \\ 
\toprule [0.2em]
Conv1 & $3 \times 3 \times 3 \times 32$ & 2 \\ \hline
\multirow{2}{*}{Conv2} & $3 \times 3 \times 32$ dw & 1 \\
& $1 \times 1 \times 32 \times 64$ & 1 \\ \hline
\multirow{2}{*}{Conv3} & $3 \times 3 \times 64$ dw & 2 \\
& $1 \times 1 \times 64 \times 128$ & 1 \\ \hline
\multirow{2}{*}{Conv4} & $3 \times 3 \times 128$ dw & 1 \\
& $1 \times 1 \times 128 \times 128$ & 1 \\ \hline
\multirow{2}{*}{Conv5} & $3 \times 3 \times 128$ dw & 2 \\
& $1 \times 1 \times 128 \times 256$ & 1 \\ \hline
\multirow{2}{*}{Conv6} & $3 \times 3 \times 256$ dw & 1 \\
& $1 \times 1 \times 256 \times 256$ & 1 \\ \hline
\multirow{2}{*}{Conv7} & $3 \times 3 \times 256$ dw & 2 \\
& $1 \times 1 \times 256 \times 512$ & 1 \\ \hline
\multirow{2}{*}{Conv8-12} & $3 \times 3 \times 512$ dw & 1 \\
& $1 \times 1 \times 512 \times 512$ & 1 \\ \hline
\multirow{2}{*}{Conv13} & $3 \times 3 \times 512$ dw & 2 \\
& $1 \times 1 \times 512 \times 1024$ & 1 \\ \hline
\multirow{4}{*}{Bottleneck-LSTM} & $3 \times 3 \times 1024$ dw & 1 \\
& $1 \times 1 \times (1024 + 256) \times 256$ & 1 \\
& $3 \times 3 \times 256$ dw & 1 \\
& $1 \times 1 \times 256 \times 1024$ & 1 \\ \hline

\multirow{3}{*}{Feature Map 1} & $1 \times 1 \times 256 \times 128$ & 1 \\
& $3 \times 3 \times 128$ dw & 2 \\
& $1 \times 1 \times 128 \times 256$ & 1 \\ \hline
\multirow{3}{*}{Feature Map 2} & $1 \times 1 \times 256 \times 64$ & 1 \\
& $3 \times 3 \times 64$ dw & 2 \\
& $1 \times 1 \times 64 \times 128$ & 1 \\ \hline
\multirow{3}{*}{Feature Map 3} & $1 \times 1 \times 128 \times 64$ & 1 \\
& $3 \times 3 \times 64$ dw & 2 \\
& $1 \times 1 \times 64 \times 128$ & 1 \\ \hline
\multirow{3}{*}{Feature Map 4} & $1 \times 1 \times 128 \times 32$ & 1 \\
& $3 \times 3 \times 32$ dw & 2 \\
& $1 \times 1 \times 32 \times 64$ & 1 \\ 
\bottomrule [0.2em]
\end{tabular}
}
\centering
\caption{Convolutional layers in one of our LSTM-SSD architectures using a single LSTM with a 256-channel state. ``dw'' denotes a depthwise convolution. Final bounding boxes are obtained by applying an additional convolution to the Bottleneck-LSTM and Feature Map layers. Note that we combine all four LSTM gate computations into a single convolution, so the LSTM computes 1024 channels of gates but outputs only 256 channels. }
\label{tab:1}
\end{table}

We are also interested in making the LSTM itself more efficient. Let $M$ and $N$ be the number of input and output channels in the LSTM respectively. Since the definition of convolutional LSTMs varies among different works \cite{ShiX, Patraucean}, we will define a standard convolutional LSTM as:
\begin{align*}
&f_t = \sigma(\leftidx{^{(M+N)}}{W}{^N}_f \star [x_t, h_{t-1}]) \\
&i_t = \sigma(\leftidx{^{(M+N)}}{W}{^N}_i \star [x_t, h_{t-1}]) \\
&o_t = \sigma(\leftidx{^{(M+N)}}{W}{^N}_o \star [x_t, h_{t-1}]) \\
&c_t = f_t \circ c_{t-1} + i_t \circ \phi(\leftidx{^{(M+N)}}{W}{^N}_c \star [x_t, h_{t-1}]) \\
&h_t = o_t \circ \phi(c_t).
\end{align*}

\begin{figure}
\centering
\subfloat{\includegraphics[width=\columnwidth]{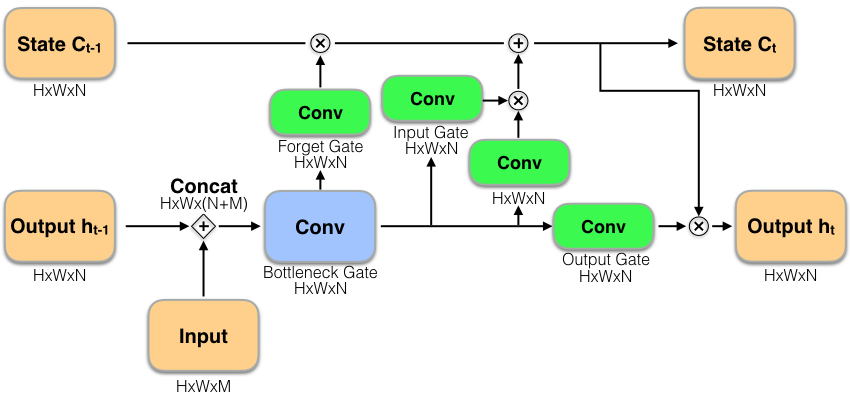}}\\
\caption{Illustration of our Bottleneck-LSTM. Note that after the bottleneck gate, all layers have only N channels.}
\label{fig:bottleneckLSTM}
\end{figure}

This LSTM takes 3D feature maps $x_t$ and $h_{t-1}$ as inputs and concatenates them channel-wise. It outputs a feature map $h_t$ and cell state $c_t$. Additionally, $\leftidx{^j}{W}{^k} \star X$ denotes a depthwise separable convolution with weights $W$, input $X$, $j$ input channels, and $k$ output channels, $\phi$ denotes the activation function, and $\circ$ denotes the Hadamard product.

Our use of depthwise separable convolutions immediately reduces the required computation by 8 to 9 times compared to previous definitions. We also choose a slightly unusual $\phi(x)=ReLU(x)$. Though ReLU activations are not commonly used in LSTMs, we find it important not to alter the bounds of the feature maps since our LSTMs are interspersed among convolutional layers.

We also wish to leverage the fact that our LSTM has substantially fewer output channels than input channels. We introduce a minor modification to the LSTM equations by first computing a bottleneck feature map with $N$ channels:
\begin{equation}
b_t = \phi(\leftidx{^{(M+N)}}{W}{^N}_b \star [x_t, h_{t-1}]). \label{eq:2} \\
\end{equation}
Then, $b_t$ replaces the inputs in all other gates as shown in Figure \ref{fig:bottleneckLSTM}. We refer to this new formulation as the Bottleneck-LSTM. The benefits of this modification are twofold. Using the bottleneck feature map decreases computation in the gates, outperforming standard LSTMs in all practical scenarios. Secondly, the Bottleneck-LSTM is deeper than the standard LSTM, which follows empirical evidence \cite{he2016deep, he2016identity} that deeper models outperform wide and shallow models.

Let the spatial dimensions of each feature map be $D_F \times D_F$, and let the dimensions of each depthwise convolutional kernel be $D_K \times D_K$. Then, the computational cost of a standard LSTM is:
\begin{equation} \label{eq:3}
4(D^2_K \cdot (M+N) \cdot D^2_F + (M+N) \cdot N \cdot D^2_F).
\end{equation}
The cost of a standard GRU is nearly identical, except with a leading coefficient of 3 instead of 4. Meanwhile, the cost of a Bottleneck-LSTM is:
\begin{equation} \label{eq:4}
\begin{split}
D^2_K \cdot (M+N) \cdot D^2_F + (M+N) \cdot N \cdot D^2_F \\
+ 4(D^2_K \cdot N \cdot D^2_F + N^2 \cdot D^2_F).
\end{split}
\end{equation}
Now, set $D_K=3$ and let $k=\frac{M}{N}$. Then, Equation \eqref{eq:3} is greater than Equation \eqref{eq:4} when $k > \frac{1}{3}$. That is, as long as our LSTM's output has less than three times as many channels as the input, it is more efficient than a standard LSTM. Since it would be extremely unusual for this condition to not hold, we claim that the Bottleneck-LSTM is more efficient in all practical situations. The Bottleneck-LSTM is also more efficient than a GRU when $k > 1$. In our network, $k=4$, and the Bottleneck-LSTM is substantially more efficient than any alternatives. One of our complete architectures is detailed in Table \ref{tab:1}.

\section{Experiments}
\subsection{Experiment Setup}
We train and evaluate on the Imagenet VID 2015 dataset. For training, we use all 3,862 videos in the Imagenet VID training set. We unroll the LSTM to 10 steps and train on sequences of 10 frames. We train our model in Tensorflow \cite{tensorflow2015-whitepaper} using RMSprop \cite{Tieleman} and asynchronous gradient descent. Like the original Mobilenet, our model can be customized to meet specific computational budgets. We present results for models with width multiplier $\alpha=1$ and $\alpha=0.5$. For the $\alpha=1$ model, we use an input resolution of $320 \times 320$ and a learning rate of $0.003$. For the $\alpha=0.5$ model, we use $256 \times 256$ input resolution and a learning rate of $0.002$.

We include hard negative mining and data augmentation as described in \cite{liu2016ssd}. We adjust the original hard negative mining approach by allowing a ratio of 10 negative examples for each positive while scaling each negative loss by $0.3$. We obtain significantly better accuracy with this modification, potentially because the original approach harshly penalizes false negatives in the groundtruth labels.

To deal with overfitting, we train the network using a two-stage process. First, we finetune the SSD network without LSTMs. Then, we freeze the weights in the base network, up to and including layer Conv13, and inject the LSTM layers for the remainder of the training.

For evaluation, we randomly select a segment of 20 consecutive frames from each video in the Imagenet VID evaluation set for a total of 11080 frames. We designate these frames as the minival set. For all results, we report the standard Imagenet VID accuracy metric, mean average precision @0.5 IOU. We also report the number of parameters and multiply-adds (MAC) as benchmarks for efficiency.

\subsection{Ablation Study}
\label{sec:4.2}
In this section, we demonstrate the individual effectiveness of each of our major design decisions.

\paragraph{Single LSTM Placement}
First, we place a single LSTM after various layers in our model. Table \ref{tab:2} confirms that placing the LSTM after feature maps results in superior performance with the Conv13 layer providing the greatest improvement, validating our claim that adding temporal awareness in the feature space is beneficial.

\begin{table}[t]
\begin{tabular}{c | c }
Placed After & mAP \\ 
\toprule [0.2em]
No LSTM (baseline) & 50.3 \\
Conv3 & 49.1 \\
Conv13 & \textbf{53.5} \\
Feature Map 1 & 51.0 \\
Feature Map 2 & 50.5 \\
Feature Map 3 & 50.8 \\
Feature Map 4 & 51.0 \\
Outputs & 51.2 \\
\bottomrule [0.2em]
\end{tabular}
\centering
\caption{Performance of our model ($\alpha=1$) with a single LSTM placed after different layers. Layer names match Table \ref{tab:1}. For the outputs experiment, we place LSTMs after all five final prediction layers.}
\label{tab:2}
\end{table}

\paragraph{Recurrent Layer Type}
Next, we compare our proposed Bottleneck-LSTM with other recurrent layer types including averaging, LSTMs, and GRUs \cite{ChoK}. For this experiment, we place a single recurrent layer after the Conv13 layer and evaluate our model for both $\alpha=1$ and $\alpha=0.5$. As a baseline, we use weighted averaging of features in consecutive frames, with a weight of $0.75$ for the current frame and $0.25$ for the previous frame. The Bottleneck-LSTM's output channel dimension is reduced according to the extended width multiplier, but all other recurrent layer types have the same input and output dimensions since they are not designed for bottlenecking. Results are shown in Table \ref{tab:3}. Our Bottleneck-LSTM is an order of magnitude more efficient than other recurrent layers while attaining comparable performance.

\begin{table}[t]
\resizebox{\columnwidth}{!}{
\begin{tabular}{c | c | c | c | c }
$\alpha$ & Type & mAP & Params (M) & MAC (M) \\ 
\toprule [0.2em]
\multirow{4}{*}{0.5} & Averaging & 37.6 & -- & -- \\
& LSTM            & 43.3 & 2.11  & 135 \\
& GRU             & \textbf{44.5} & 1.59 & 102 \\
& Bottleneck-LSTM & 42.8 & \textbf{0.15} & \textbf{5.6} \\ \hline
\multirow{4}{*}{1.0} & Averaging & 50.7 & -- & -- \\
& LSTM            & 53.5 & 8.41 & 840 \\
& GRU             & \textbf{54.0} & 6.33 & 632 \\
& Bottleneck-LSTM & 53.5 & \textbf{0.60} & \textbf{34} \\
\bottomrule [0.2em]
\end{tabular}
}
\centering
\caption{Performance of different recurrent layer types. The parameters and multiply-adds columns are computed for the recurrent layer only.}
\label{tab:3}
\end{table}

\paragraph{Bottleneck Dimension}

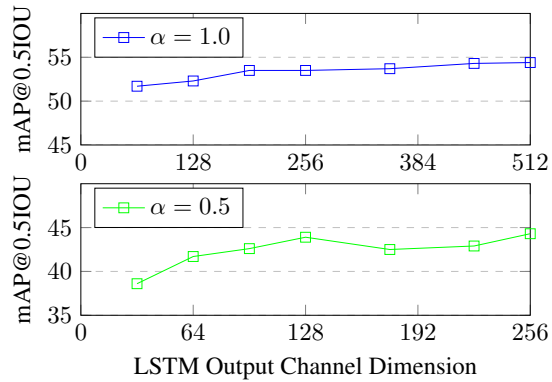
\begin{figure}
\centering
\subfloat{\resizebox{0.9\columnwidth}{!}{
\begin{tikzpicture}
\begin{axis}[
    ylabel={mAP@0.5IOU},
    xmin=0, xmax=512,
    ymin=45, ymax=60,
    y=0.5cm/4,
    x=0.5cm/40,
    xtick={0, 128, 256, 384, 512},
    ytick={45, 50, 55},
    legend pos=north west,
    ymajorgrids=true,
    grid style=dashed,
]
 
\addplot[
    color=blue,
    mark=square,
    ]
    coordinates {
    (512,54.4)(448,54.3)(352,53.7)(256,53.5)(192,53.5)(128,52.3)(64,51.7)
    };
    \legend{$\alpha=1.0$}
\end{axis}
\end{tikzpicture}
}}\\
\vspace{-1.5em}
\subfloat{\resizebox{0.9\columnwidth}{!}{
\begin{tikzpicture}
\begin{axis}[
    xlabel={LSTM Output Channel Dimension},
    ylabel={mAP@0.5IOU},
    xmin=0, xmax=256,
    ymin=35, ymax=50,
    y=0.5cm/4,
    x=0.5cm/20,
    xtick={0, 64, 128, 192, 256},
    ytick={35, 40, 45},
    legend pos=north west,
    ymajorgrids=true,
    grid style=dashed,
]
\addplot[
    color=green,
    mark=square,
    ]
    coordinates {
    (256,44.3)(224,42.9)(176,42.5)(128,43.9)(96,42.6)(64,41.7)(32,38.6)
    };
    \legend{$\alpha=0.5$}
\end{axis}
\end{tikzpicture}
}}
\centering
\caption{Model Performance vs. LSTM Output Channels. This figure shows model performance as a function of the LSTM's output channel dimension.}
\label{fig:4}
\end{figure}

We further analyze the effect of the LSTM output channel dimension on accuracy, shown in Figure \ref{fig:4}. A single Bottleneck-LSTM is placed after the Conv13 layer in each experiment. Accuracy remains near-constant up to $\alpha_{lstm}=0.25\alpha$, then drops off. This supports our use of the extended width multiplier.

\paragraph{Multiple LSTM Placement Strategies}

\begin{table*}[t]
\centering
\begin{tabular}{c  c  c  c  c | c | c | c | c | c | c }
\multicolumn{5}{c|}{Place Bottleneck-LSTMs After} & \multicolumn{2}{c|}{mAP} & \multicolumn{2}{c|}{Params (M)} & \multicolumn{2}{c}{MAC (B)} \\
\toprule [0.2em]
Conv13 & FM1 & FM 2 & FM 3 & FM 4 & $\alpha=0.5$ & $\alpha=1.0$ & $\alpha=0.5$ & $\alpha=1.0$ & $\alpha=0.5$ & $\alpha=1.0$ \\ \hline
\checkmark & & & & & 42.8 & 53.5 & \textbf{0.81} & \textbf{3.01} & \textbf{0.19} & \textbf{1.12} \\
\checkmark \checkmark & & & & & 42.9 & 53.3 & 0.91 & 3.41 & 0.20 & 1.16 \\
\checkmark & \checkmark & & & & 43.3 & 53.7 & 0.82 & 3.07 & 0.19 & 1.13  \\
\checkmark & \checkmark & \checkmark & & & 43.5 & 54.0 & 0.82 & 3.11 & 0.19 & 1.13  \\
\checkmark & \checkmark & \checkmark & \checkmark & & \textbf{43.8} & \textbf{54.4} & 0.85 & 3.21 & 0.19 & 1.13 \\
\checkmark & \checkmark & \checkmark & \checkmark & \checkmark & \textbf{43.8} & \textbf{54.4} & 0.86 & 3.24 & 0.19 & 1.13 \\
\bottomrule [0.2em]
\end{tabular}
\caption{Performance of our architecture with multiple Bottleneck-LSTMs. Layer names match Table \ref{tab:1}, where FM stands for Feature Map. \checkmark denotes a single LSTM placed after the specified layer, while \checkmark \checkmark denotes two LSTMs placed.}
\label{tab:multi_placement}
\end{table*}

Our framework naturally generalizes to multiple LSTMs. In SSD, each feature map represents features at a certain scale. We investigate the benefit of incorporating multiple LSTMs to refine feature maps at different scales.  In Table \ref{tab:multi_placement}, we evaluate different strategies for incorporating multiple LSTMs. In this experiment, we incrementally add more LSTM layers to the network. Due to difficulties in training multiple LSTMs simultaneously, we finetune from previous checkpoints while progressively adding layers. Placing LSTMs after feature maps at different scales results in a slight performance improvement and nearly no change in computational cost due to the small dimensions of later feature maps and the efficiency of our LSTM layers. However, stacking two LSTMs after the same feature map is not beneficial. For the last two feature maps (FM3, FM4), we do not further bottleneck the LSTM output channels because the channel dimension is already very small. We use the model with LSTMs placed after all feature maps as our final model.

\subsection{Comparison With Other Architectures}

\begin{table}[t]
\resizebox{\columnwidth}{!}{
\begin{tabular}{c | c | c | c }
Model & mAP & Params (M) & MAC (B) \\ 
\toprule [0.2em]
Resnet-101 Faster-RCNN & 62.5 & 63.2 & 143.37 \\
Inception-SSD & 56.3 & 13.7 & 3.79 \\
\hline
Mobilenet-SSD ($\alpha = 1$) & 50.3 & 4.20 & 1.20 \\
Ours ($\alpha = 1$) & \textbf{54.4} & \textbf{3.24} & \textbf{1.13} \\ \hline
Mobilenet-SSD ($\alpha = 0.5$) & 40.0 & 1.17 & 0.20 \\
Ours ($\alpha = 0.5$) & \textbf{43.8} & \textbf{0.86} & \textbf{0.19} \\
\bottomrule [0.2em]
\end{tabular}
}
\centering
\caption{Final results on our Imagenet VID minival set.}
\label{tab:5}
\end{table}

\begin{figure*}
\centering
\subfloat{\includegraphics[width=.16\textwidth,height=.215\textwidth]{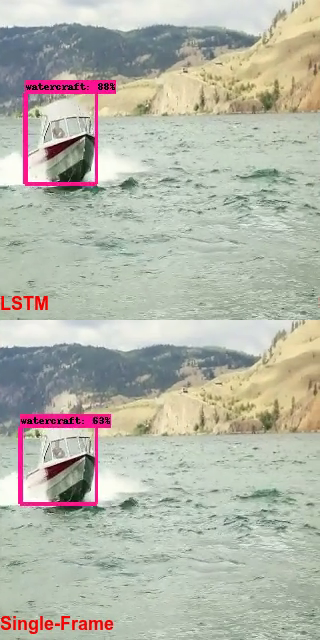}}%
\subfloat{\includegraphics[width=.16\textwidth,height=.215\textwidth]{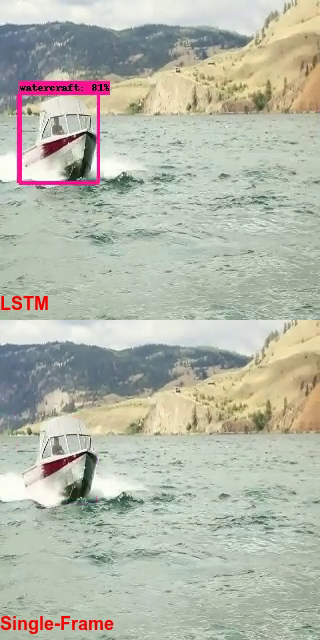}}%
\subfloat{\includegraphics[width=.16\textwidth,height=.215\textwidth]{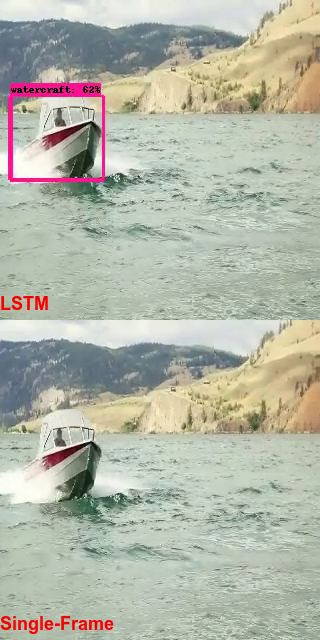}}%
\subfloat{\includegraphics[width=.16\textwidth,height=.215\textwidth]{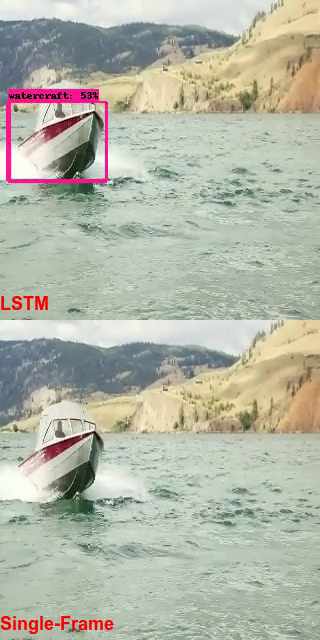}}%
\subfloat{\includegraphics[width=.16\textwidth,height=.215\textwidth]{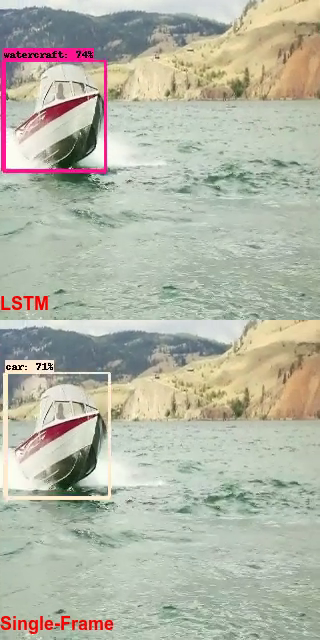}}%
\subfloat{\includegraphics[width=.16\textwidth,height=.215\textwidth]{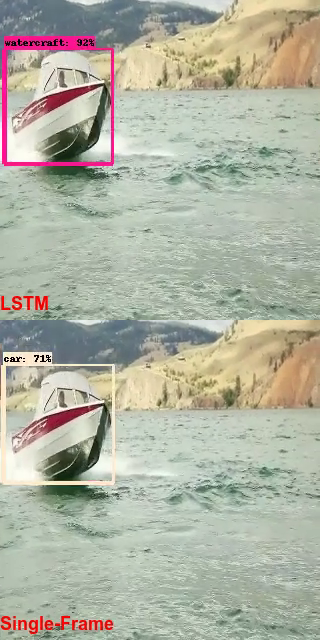}}\\\vspace{-.7em}
\subfloat{\includegraphics[width=.16\textwidth,height=.215\textwidth]{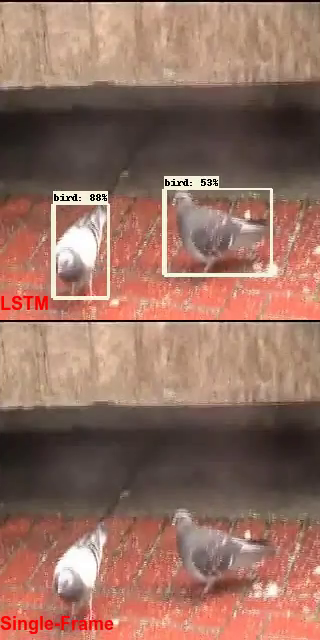}}%
\subfloat{\includegraphics[width=.16\textwidth,height=.215\textwidth]{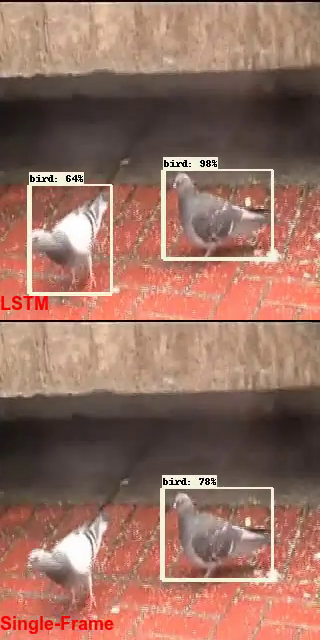}}%
\subfloat{\includegraphics[width=.16\textwidth,height=.215\textwidth]{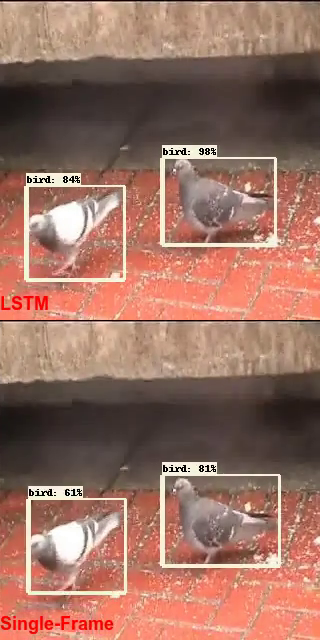}}%
\subfloat{\includegraphics[width=.16\textwidth,height=.215\textwidth]{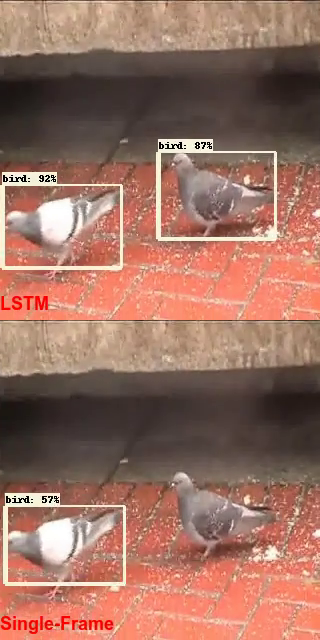}}%
\subfloat{\includegraphics[width=.16\textwidth,height=.215\textwidth]{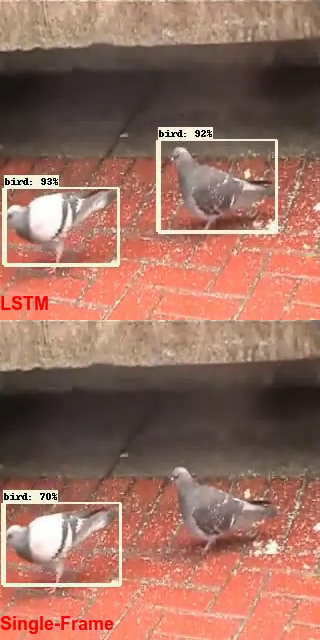}}%
\subfloat{\includegraphics[width=.16\textwidth,height=.215\textwidth]{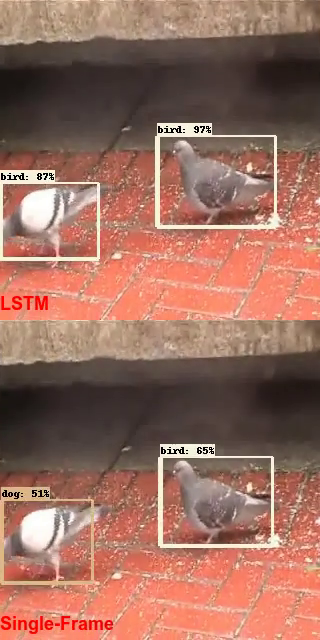}}\\\vspace{-.7em}
\subfloat{\includegraphics[width=.16\textwidth,height=.215\textwidth]{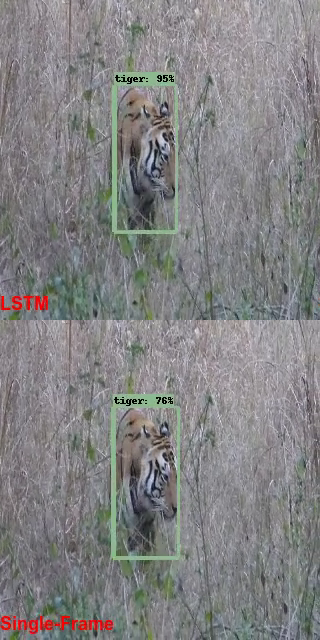}}%
\subfloat{\includegraphics[width=.16\textwidth,height=.215\textwidth]{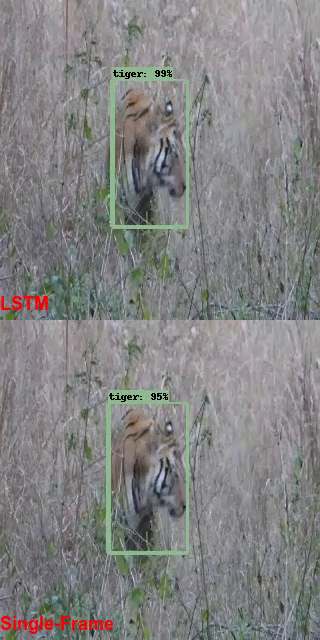}}%
\subfloat{\includegraphics[width=.16\textwidth,height=.215\textwidth]{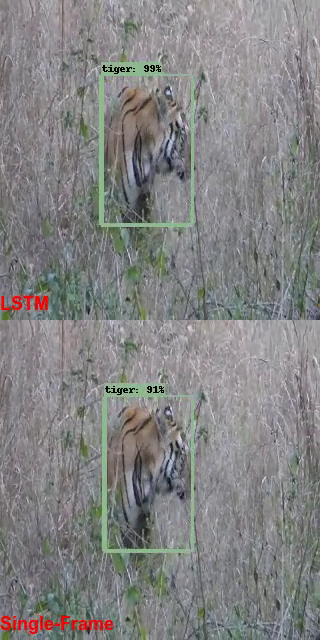}}%
\subfloat{\includegraphics[width=.16\textwidth,height=.215\textwidth]{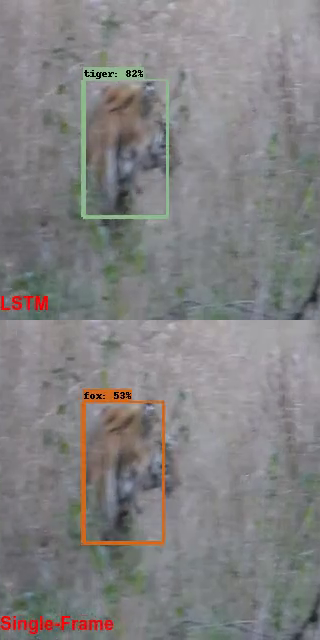}}%
\subfloat{\includegraphics[width=.16\textwidth,height=.215\textwidth]{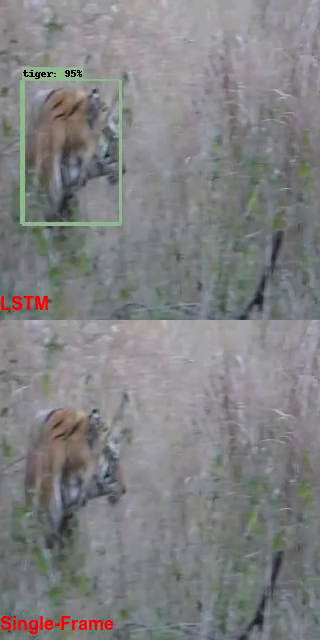}}%
\subfloat{\includegraphics[width=.16\textwidth,height=.215\textwidth]{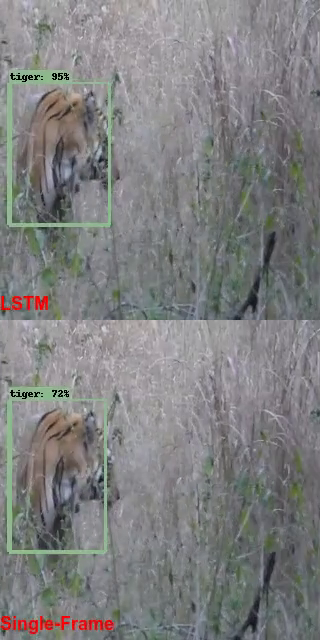}}\\\vspace{-.7em}
\subfloat{\includegraphics[width=.16\textwidth,height=.215\textwidth]{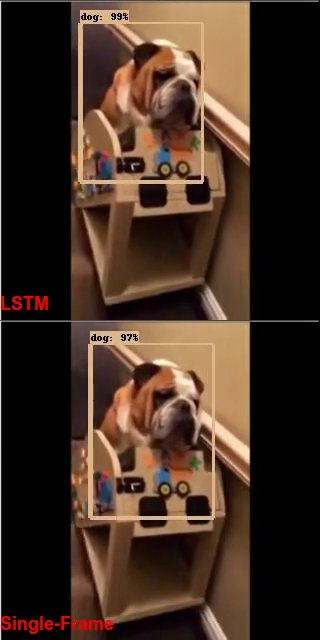}}
\subfloat{\includegraphics[width=.16\textwidth,height=.215\textwidth]{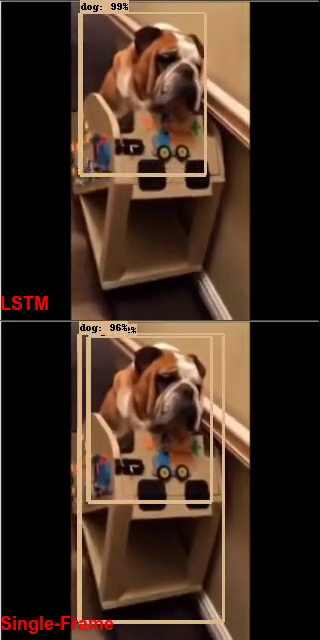}}%
\subfloat{\includegraphics[width=.16\textwidth,height=.215\textwidth]{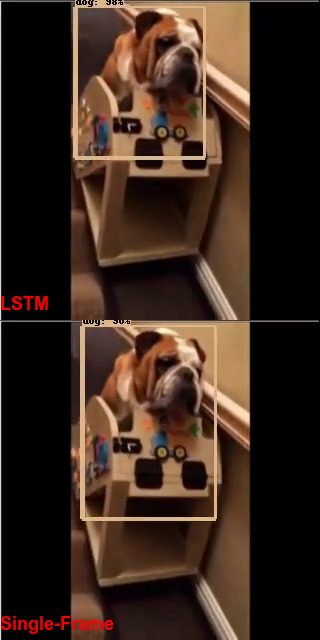}}%
\subfloat{\includegraphics[width=.16\textwidth,height=.215\textwidth]{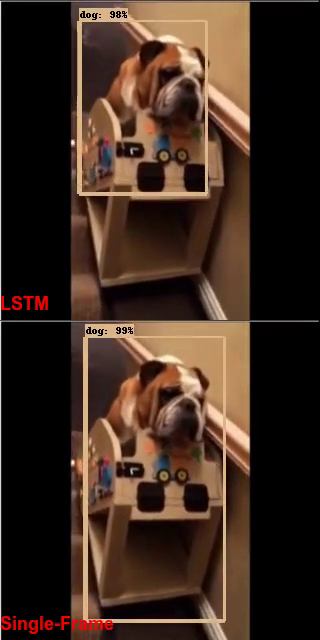}}%
\subfloat{\includegraphics[width=.16\textwidth,height=.215\textwidth]{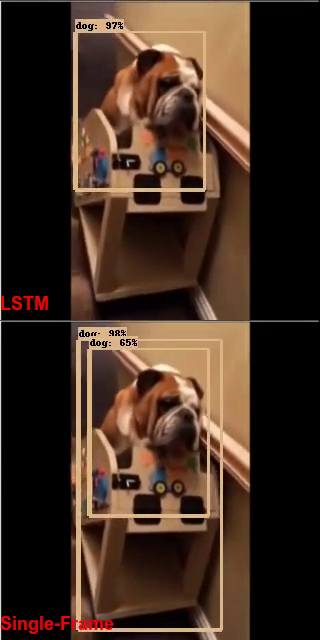}}%
\subfloat{\includegraphics[width=.16\textwidth,height=.215\textwidth]{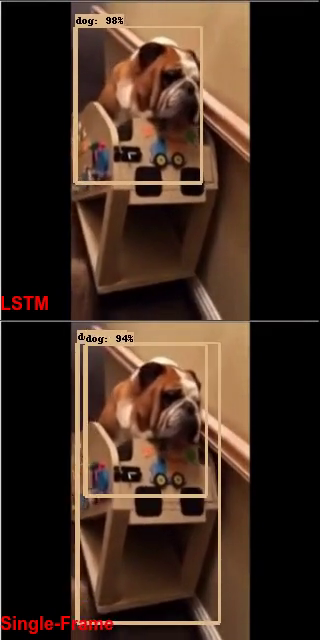}}%
\caption{Example clips in Imagenet VID minival set where our model ($\alpha=1$) outperforms an analogous single-frame Mobilenet-SSD model ($\alpha=1$). Our network uses temporal context to provide significantly more stable detections across frames. The upper row of each sequence corresponds to our model, and the lower row corresponds to Mobilenet-SSD.}
\label{fig:visualization}
\end{figure*}

In Table \ref{tab:5}, we compare our final model against state-of-the-art single-image detection frameworks. All baselines are trained with the open source Tensorflow Object Detection API \cite{detection-api}. Among these methods, only the Mobilenet-SSD variants and our method can run in real-time on mobile devices, and our method outperforms Mobilenet-SSD on all metrics. Qualitative differences are shown in Figure \ref{fig:visualization}. We also include performance-focused architectures which are much more computationally expensive. Our method approaches the accuracy of the Inception-SSD network at a fraction of the cost. Additionally, we include the parameters and MAC of a Resnet Faster-RCNN to highlight the vast computational difference between our method and non-mobile video object detection methods, which generally include similarly expensive base networks.

\subsection{Robustness Evaluation}
\begin{table}[t]
\resizebox{\columnwidth}{!}{
\begin{tabular}{c | c | c | c }
Model & \multicolumn{3}{c}{mAP} \\ 
\toprule [0.2em]
& $p = 0.25$ &  $p = 0.50$ & $p = 0.75$ \\ 
\hline
Inception-SSD & 43.3 & 34.6 & 25.9 \\
Mobilenet-SSD ($\alpha = 1$) & 43.0 & 33.8 & 24.6 \\
Ours ($\alpha = 1$) & \textbf{49.1} & \textbf{42.4} & \textbf{33.3} \\ 
\hline
Mobilenet-SSD ($\alpha = 0.5$) & 33.2 & 26.0 & 19.3 \\
Ours ($\alpha = 0.5$) & \textbf{39.8} & \textbf{33.9} & \textbf{25.6} \\
\bottomrule [0.2em]
\end{tabular}
}
\centering
\caption{Model performance after randomly occluding a fraction $p$ of the bounding boxes.}
\label{tab:6}
\end{table}
We test the robustness of our method to input noise by creating artificial occlusions in each video. We generate these occlusions as follows: for each groundtruth bounding box, we assign a probability $p$ of occluding the bounding box. For each occluded bounding box of dimension $H \times W$, we zero out all pixels in a randomly selected rectangular region of size between $\frac{H}{2} \times \frac{W}{2}$ and $\frac{3H}{4} \times \frac{3W}{4}$ within that bounding box. Results for this experiment are reported in Table \ref{tab:6}. All methods are evaluated on the same occlusions, and no method is trained on these occlusions prior to testing. Our method outperforms all single-frame SSD methods on this noisy data, demonstrating that our network has learned the temporally continuous nature of videos and uses temporal cues to achieve robustness against noise.

\begin{table}[t]
    \centering
    \resizebox{\columnwidth}{!}{
    \begin{tabular}{c|cc}
        Model & big core (ms) & LITTLE core (ms)\\
        \toprule [0.2em]
        Inception-SSD & 1020 & 2170 \\
        Mobilenet-SSD ($\alpha=1$) & 357 & 794 \\
        Ours ($\alpha=1$) & \textbf{322} & \textbf{665} \\
        \hline
        Mobilenet-SSD ($\alpha=0.5$) & 72 & 157 \\
        Ours ($\alpha=0.5$) & \textbf{65} & \textbf{140} \\
        \bottomrule [0.2em]
    \end{tabular}
    }
    \caption{Runtime comparison of single-frame detection models and our LSTM-SSD model on both the big and LITTLE core of a Qualcomm Snapdragon 835 on a Pixel 2.}
    \label{tab:runtime}
\end{table}

\subsection{Mobile Runtime}
We evaluate our models against baselines on the latest Pixel 2 phone with Qualcomm Snapdragon 835. Runtime is measured on both Snapdragon 835 big and LITTLE core, with single-threaded inference using a custom on-device implementation of Tensorflow \cite{tensorflow2015-whitepaper}. Table \ref{tab:runtime} shows that our model outperforms all baselines. Notably, our $\alpha=0.5$ model achieves real-time speed of 15 FPS on the big core.

\section{Conclusion}
We introduce a novel framework for mobile objection detection in videos based on unifying mobile SSD frameworks and recurrent networks into a single temporally-aware architecture. We propose an array of modifications that allow our model to be faster and more lightweight than mobile-focused single-frame models despite having a more complex architecture. We proceed to examine each of our design decisions individually, and empirically show that our modifications allow our network to be more efficient with minimal decrease in performance. We also demonstrate that our network is sufficiently fast to run in real-time on mobile devices. Finally, we show that our method outperforms comparable state-of-the-art single-frame models to indicate that our network benefits from temporal cues in videos.

\paragraph{Acknowledgements} We gratefully appreciate the help and discussions in the course of this work from our colleagues: Yuning Chai, Matthew Tang, Benoit Jacob, Liang-Chieh Chen, Susanna Ricco and Bryan Seybold.

\clearpage
{\small
\bibliographystyle{ieee}
\bibliography{paper}
}
\end{document}